\documentclass[lettersize,journal]{IEEEtran}
\usepackage{amsmath,amsfonts}
\usepackage{algorithmic}
\usepackage{algorithm}
\usepackage{array}
\usepackage[caption=false,font=normalsize,labelfont=sf,textfont=sf]{subfig}
\usepackage{textcomp}
\usepackage{stfloats}
\usepackage{url}
\usepackage{verbatim}
\usepackage{graphicx}
\usepackage{cite}
\begin{document}

\title{CLAReSNet: When Convolution Meets Latent Attention for Hyperspectral Image Classification}

\author{Asmit Bandyopadhyay, Anindita Das Bhattacharjee, Rakesh Das
\thanks{Asmit Bandyopadhyay and Anindita Das Bhattacharjee are with Institute of Engineering and Management (IEM), University of Engineering and Management Kolkata, IEM Centre of Excellence for InnovAI, Department of Computer Science \& Engineering, Kolkata, West Bengal, India.

Rakesh Das is with Institute of Engineering and Management (IEM), University of Engineering and Management Kolkata, Department of Computer Science \& Engineering, Kolkata, West Bengal, India.}
\thanks{All the datasets and codes used in this research article are available at https://github.com/Bandyopadhyay-Asmit/CLAReSNet}}

\markboth{Bandyopadhyay \MakeLowercase{\textit{et al.}}: CLAR\MakeLowercase{e}SN\MakeLowercase{et} for HSI classification}%
{Journal of \LaTeX\ Class Files,~Vol.~14, No.~8, August~2021}


\maketitle

\begin{abstract}
Hyperspectral image (HSI) classification faces critical challenges, including high spectral dimensionality, complex spectral-spatial correlations, and limited training samples with severe class imbalance. While CNNs excel at local feature extraction and transformers capture long-range dependencies, their isolated application yields suboptimal results due to quadratic complexity and insufficient inductive biases. We propose CLAReSNet (Convolutional Latent Attention Residual Spectral Network), a hybrid architecture that integrates multi-scale convolutional extraction with transformer-style attention via an adaptive latent bottleneck. The model employs a multi-scale convolutional stem with deep residual blocks and an enhanced Convolutional Block Attention Module for hierarchical spatial features, followed by spectral encoder layers combining bidirectional RNNs (LSTM/GRU) with Multi-Scale Spectral Latent Attention (MSLA). MSLA reduces complexity from $\mathcal{O}(T^2D)$ to $\mathcal{O}(T\log(T)D)$ by adaptive latent token allocation (8-64 tokens) that scales logarithmically with the sequence length. Hierarchical cross-attention fusion dynamically aggregates multi-level representations for robust classification. Experiments conducted on the Indian Pines and Salinas datasets show state-of-the-art performance, achieving overall accuracies of 99.71\% and 99.96\%, significantly surpassing HybridSN, SSRN, and SpectralFormer. The learned embeddings exhibit superior inter-class separability and compact intra-class clustering, validating CLAReSNet's effectiveness under severe class imbalance.

\end{abstract}


\section{Introduction}
\IEEEPARstart{H}{yperspectral} imaging has emerged as a powerful remote sensing technology, capturing hundreds of narrow and contiguous spectral bands across the electromagnetic spectrum to provide fine grained material identification capabilities~\cite{ref1}. Hyperspectral image (HSI) classification, the task of assigning class labels to individual pixels based on their spectral-spatial information, is fundamental to numerous applications, including land-use monitoring, mineral exploration, agricultural assessment, and environmental management~\cite{ref2}. However, HSI classification faces three critical challenges that significantly impact model performance and generalization. First, the \textbf{high dimensionality} inherent in hyperspectral data, with hundreds of spectral bands, introduces the curse of dimensionality, rendering traditional machine learning approaches computationally intractable and susceptible to overfitting~\cite{ref3}. Second, \textbf{complex spectral-spatial correlations} between adjacent bands create strong nonlinear dependencies that demand sophisticated feature extraction mechanisms capable of modeling both local spectral patterns and global spatial relationships~\cite{ref4}. Third, \textbf{variable sequence lengths} across different sensors and acquisition protocols necessitate adaptive architectures that can flexibly handle disparate input dimensions without performance degradation~\cite{ref5}.

Historically, hyperspectral classification relied on handcrafted feature extraction combined with classical machine learning algorithms. Support Vector Machines (SVMs) with kernel methods dominated the field due to their effectiveness in high-dimensional spaces~\cite{ref6}. More recently, ensemble methods such as \textbf{Random Forest} and gradient boosting techniques including \textbf{XGBoost} have demonstrated competitive performance when combined with dimensionality reduction strategies like Principal Component Analysis (PCA) and feature selection methods~\cite{ref26,ref7}. While these approaches remain interpretable and computationally efficient, they require extensive feature engineering and struggle to capture the intricate nonlinear relationships present in hyperspectral data. The advent of deep learning revolutionised HSI classification through convolutional neural networks' ability to extract hierarchical features automatically. The seminal \textbf{3D-CNN} approaches jointly extract spectral-spatial features by treating contiguous spectral bands as a 3D volume~\cite{ref8}. Building on this foundation, \textbf{SSRN (Spectral Spatial Residual Network)}~\cite{ref23} introduced residual learning to improve information flow through deeper architectures, while \textbf{HybridSN}~\cite{ref9} proposed a hybrid 3D-2D CNN strategy that combines the spectral-spatial learning capability of 3D convolutions with the computational efficiency and abstract spatial modeling of 2D convolutions. These CNN-based methods excel at capturing local context but inherently suffer from limited receptive fields, potentially missing long-range spectral-spatial dependencies and global discriminative patterns essential for accurate classification~\cite{ref10}.

Recent advances have introduced transformer architectures to overcome CNN limitations. \textbf{SpectralFormer}~\cite{ref11} pioneered transformer-based HSI classification by treating spectral bands as a sequence of tokens, enabling transformers to capture long-range spectral dependencies through self-attention mechanisms. The architecture leverages cross-layer adaptive fusion to preserve critical information across layers~\cite{ref11}. However, pure transformer models often struggle with computational efficiency on high-resolution spatial patches and may neglect local feature details critical for fine grained discrimination~\cite{ref25}. Recognizing the complementary strengths of CNNs and transformers, recent works have explored fusion strategies~\cite{ref25}. \textbf{Cross-attention fusion networks} integrate CNN branches for local feature extraction with transformer branches for global context modeling, using cross-attention modules to enable bidirectional information flow between modalities~\cite{ref12}. Multi-scale CNN-Transformer hybrid architectures enhance feature representation by capturing features at different spatial scales and fusing them through learned attention mechanisms~\cite{ref13}. Additionally, \textbf{attention bottleneck mechanisms} have shown promise in multimodal learning, where information from different sources is forced to pass through compressed latent representations, requiring models to identify and condense the most relevant information~\cite{ref14}. These innovations demonstrate that strategic fusion of CNNs and transformers with adaptive latent representations can substantially improve both accuracy and computational efficiency. Recent research emphasizes the importance of latent space compression and adaptive token management for handling variable length sequences in HSI data~\cite{ref27, ref28}. Methods incorporating learnable latent bottlenecks enable dynamic adjustment of model capacity based on input sequence length, reducing computational overhead while maintaining representational expressiveness~\cite{ref28}. Positional encoding strategies combining sinusoidal and learnable embeddings have proven effective for encoding spectral band positions, enhancing the model's ability to capture sequential dependencies in spectral signatures~\cite{ref15}.

We introduce \textbf{CLAReSNet (Convolutional Latent Attention Residual Spectral Network)}, a deep learning architecture designed for hyperspectral image classification that integrates convolutional neural networks with transformer-style attention mechanisms through a novel latent bottleneck approach. The model directly addresses the key challenges of hyperspectral data by combining three synergistic components: (1) a CNN-based spatial feature extractor that captures local spectral-spatial patterns with multi-scale convolutions, residual blocks and channel-spatial attention mechanisms, (2) a multi-scale spectral latent attention module featuring adaptive latent tokens that compress information bottlenecks based on input sequence length, and (3) stacked spectral encoder layers employing recurrent networks (LSTM/GRU) alongside latent attention to model complex spectral-spatial correlations. The architecture's innovation lies in its latent bottleneck design, which adaptively manages the number of latent tokens through logarithmic scaling relative to sequence length, effectively addressing the curse of dimensionality while preserving discriminative information. Hybrid positional encoding and residual connections maintain information integrity throughout deep network propagation.

\section{Problem Definition}

Hyperspectral image classification aims to assign each pixel in a hyperspectral image to one of \(C\) predefined classes based on its spectral signature and spatial context. Formally, given a hyperspectral image \(\mathbf{I} \in \mathbb{R}^{M \times N \times B}\), where \(M\) and \(N\) are the spatial dimensions and \(B\) is the number of spectral bands, and a ground truth label map \(\mathbf{L} \in \{0, 1, \dots, C\}^{M \times N}\) (with 0 indicating unlabeled pixels), the task is to learn a function \(f: \mathbb{R}^{P \times P \times B} \to \{1, \dots, C\}\) that maps a local patch \(\mathbf{X} \in \mathbb{R}^{P \times P \times B}\) centered at a pixel to its class label, where \(P\) is the patch size (e.g., 11).

To handle high dimensionality, Principal Component Analysis (PCA) is applied to reduce the spectral bands to \(T\) components: \(\mathbf{I}' = \text{PCA}(\mathbf{I}) \in \mathbb{R}^{M \times N \times T}\), and patches are extracted as \(\mathbf{X} \in \mathbb{R}^{T \times P \times P}\), treating spectral bands as the sequence dimension.

The model is trained to minimize the cross-entropy loss:
\[
\mathcal{L} = -\frac{1}{N_s} \sum_{i=1}^{N_s} \sum_{c=1}^{C} y_{i,c} \log(\hat{y}_{i,c}),
\]
where \(N_s\) is the number of samples, \(\mathbf{y}_i\) is the one-hot encoded ground truth label for sample \(i\), and \(\hat{\mathbf{y}}_i = \text{softmax}(f(\mathbf{X}_i))\) is the predicted probability distribution.

\section{Experimental Setup}

\subsection{Dateset Description}

Experiments are conducted on two benchmark hyperspectral remote sensing datasets: Indian Pines and Salinas. The Indian Pines dataset, captured by the AVIRIS sensor over Northwestern Indiana, has a spatial resolution of $145 \times 145$ pixels with 224 spectral bands (0.4--2.5 $\mu$m). After removing 24 water absorption bands ([104--108], [150--163], 220), 200 bands remain. The scene, mainly agricultural (two-thirds) and forest/natural vegetation (one-third), includes highways, a rail line, low-density housing, built structures, and smaller roads. Captured in June, it features early-stage crops (corn, soybeans) with $<5\%$ coverage. The ground truth has 16 non-mutually exclusive classes (Table~\ref{tab:indian_pines_classes}). The Salinas dataset, collected by AVIRIS over Salinas Valley, California, has a 3.7-meter resolution with 512$\times$217 pixels and 224 spectral bands (0.4--2.5 $\mu$m). After discarding 20 water absorption bands ([108--112], [154--167], 224), 204 bands remain. It includes vegetables, bare soils, and vineyard fields, with ground truth in 16 classes (Table~\ref{tab:salinas_classes}). Both datasets present significant challenges due to limited labeled samples, high spectral dimensionality, and substantial class imbalance.

\begin{table}[h!]
\caption{Ground Truth Classes and Sample Counts for the Indian Pines Dataset\label{tab:indian_pines_classes}}
\centering
\begin{tabular}{clc}
\hline
\textbf{Class} & \textbf{Class Name} & \textbf{Samples} \\
\hline
1 & Alfalfa & 46 \\
2 & Corn-notill & 1428 \\
3 & Corn-mintill & 830 \\
4 & Corn & 237 \\
5 & Grass-pasture & 483 \\
6 & Grass-trees & 730 \\
7 & Grass-pasture-mowed & 28 \\
8 & Hay-windrowed & 478 \\
9 & Oats & 20 \\
10 & Soybean-notill & 972 \\
11 & Soybean-mintill & 2455 \\
12 & Soybean-clean & 593 \\
13 & Wheat & 205 \\
14 & Woods & 1265 \\
15 & Buildings-Grass-Trees-Drives & 386 \\
16 & Stone-Steel-Towers & 93 \\
\hline
\end{tabular}
\end{table}

\begin{figure*}[h!]
\centering
\begin{tabular}{cc}
\includegraphics[height=2.7cm]{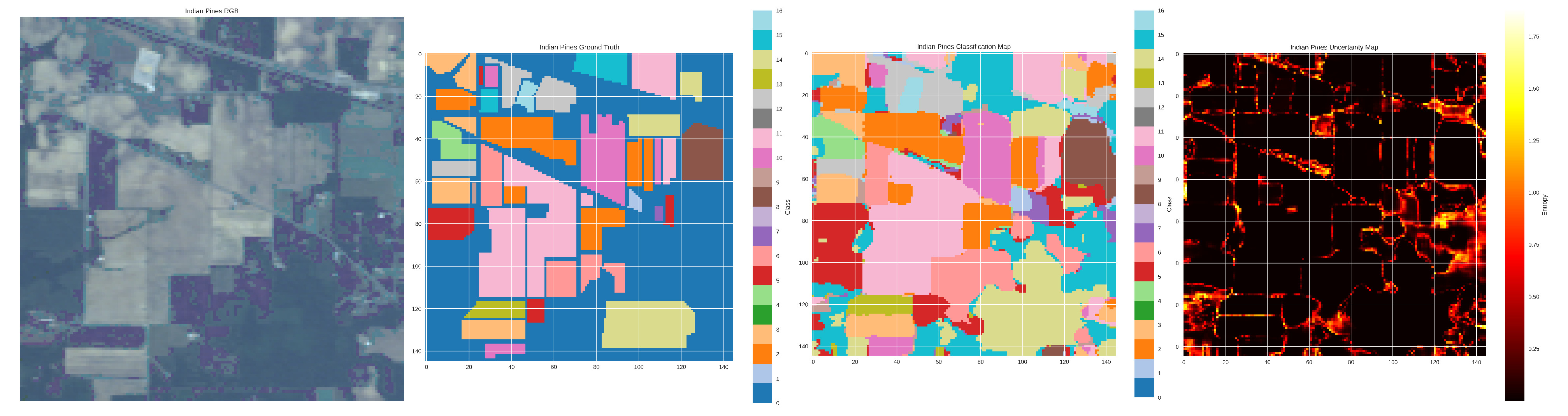} &
\includegraphics[height=3cm]{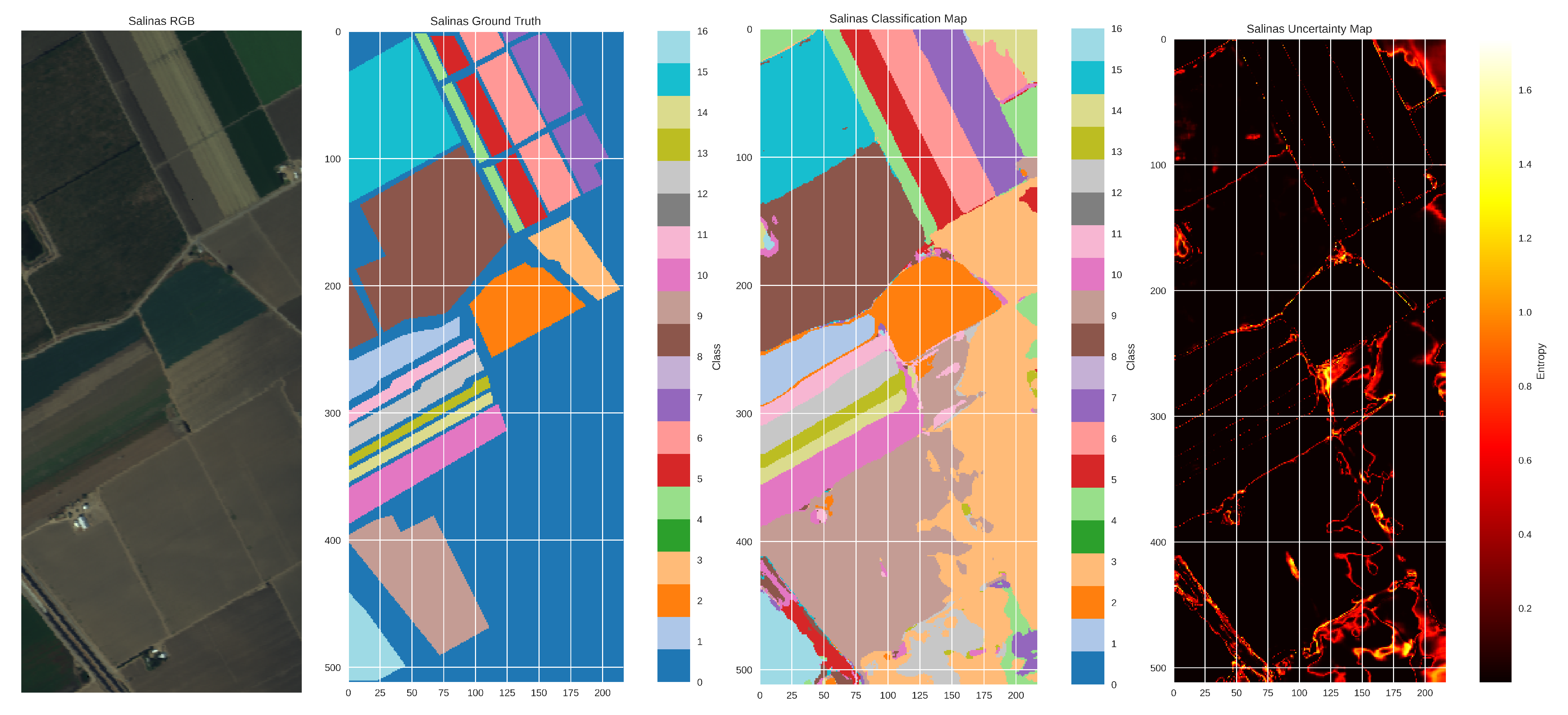} \\
(a) Indian Pines & (b) Salinas
\end{tabular}
\caption{\label{fig:datasets} (a) Indian Pines (RGB Image, Ground Truth, Classification Map, Uncertainty Map) and (b) Salinas (RGB Image, Ground Truth, Classification Map, Uncertainty Map).}
\end{figure*}

\begin{table}[h!]
\caption{Ground Truth Classes and Sample Counts for the Salinas Dataset\label{tab:salinas_classes}}
\centering
\begin{tabular}{clc}
\hline
\textbf{Class} & \textbf{Class Name} & \textbf{Samples} \\
\hline
1 & Broccoli green weeds 1 & 2009 \\
2 & Broccoli green weeds 2 & 3726 \\
3 & Fallow & 1976 \\
4 & Fallow rough plow & 1394 \\
5 & Fallow smooth & 2678 \\
6 & Stubble & 3959 \\
7 & Celery & 3579 \\
8 & Grapes untrained & 11271 \\
9 & Soil vineyard develop & 6203 \\
10 & Corn senesced green weeds & 3278 \\
11 & Lettuce romaine 4wk & 1068 \\
12 & Lettuce romaine 5wk & 1927 \\
13 & Lettuce romaine 6wk & 916 \\
14 & Lettuce romaine 7wk & 1070 \\
15 & Vineyard untrained & 7268 \\
16 & Vineyard vertical trellis & 1807 \\
\hline
\end{tabular}
\end{table}

\subsection{Data Preprocessing}

Prior to model training, both datasets are reshaped into 3D cubes (145$\times$145$\times$200 for Indian Pines; 512$\times$217$\times$204 for Salinas) and reduced to 30 components via PCA, yielding 145$\times$145$\times$30 and 512$\times$217$\times$30, respectively. For complete training, labeled pixels are split into training (72\%), validation (8\%), and testing (20\%) sets using stratified sampling. 
Patches of 11$\times$11 are extracted around labeled pixels with reflection padding for boundaries. Data augmentation is applied during training to enhance model generalization. Augmentation strategies include additive Gaussian noise (std 0.05, 50\% probability), random rotations (90°, 180°, 270°, 50\% probability), and random flips (horizontal/vertical, 50\% probability).

\subsection{Training Configuration \& Environment}
The CLAReSNet model is optimized using the AdamW~\cite{ref24} optimizer with an initial learning rate of 1e-4 and weight decay coefficient of 1e-2 to prevent overfitting. The categorical cross-entropy loss function is employed as the training objective. Training proceeds for 40 epochs with early stopping based on validation accuracy. The model checkpoint achieving the highest validation accuracy is retained for subsequent testing and analysis.

All experiments are conducted on NVIDIA P100 GPUs with 16 GB memory using the PyTorch deep learning framework within the Kaggle environment. Training employs a batch size of 16 to balance memory constraints and gradient estimate quality, while validation and testing utilize a batch size of 32 for computational efficiency. 

\subsection{Evaluation Metrics \& Visualization}

Model performance is assessed using a comprehensive suite of metrics, including overall accuracy (OA), balanced accuracy, Cohen's kappa (k), Matthews correlation coefficient (MCC), and adjusted Rand index (ARI). The average pairwise Euclidean distance is computed between class centroids in the learned feature space.
Visualizations include t-SNE feature space projections, full-scene classification maps, pixel-wise uncertainty maps (prediction entropy), and precision-recall curves. These provide comprehensive insight into model behaviour, learned representations, and spatial prediction patterns.

\section{Methodology}
\begin{figure*}[h!]
\centering
\includegraphics[height=7.5cm]{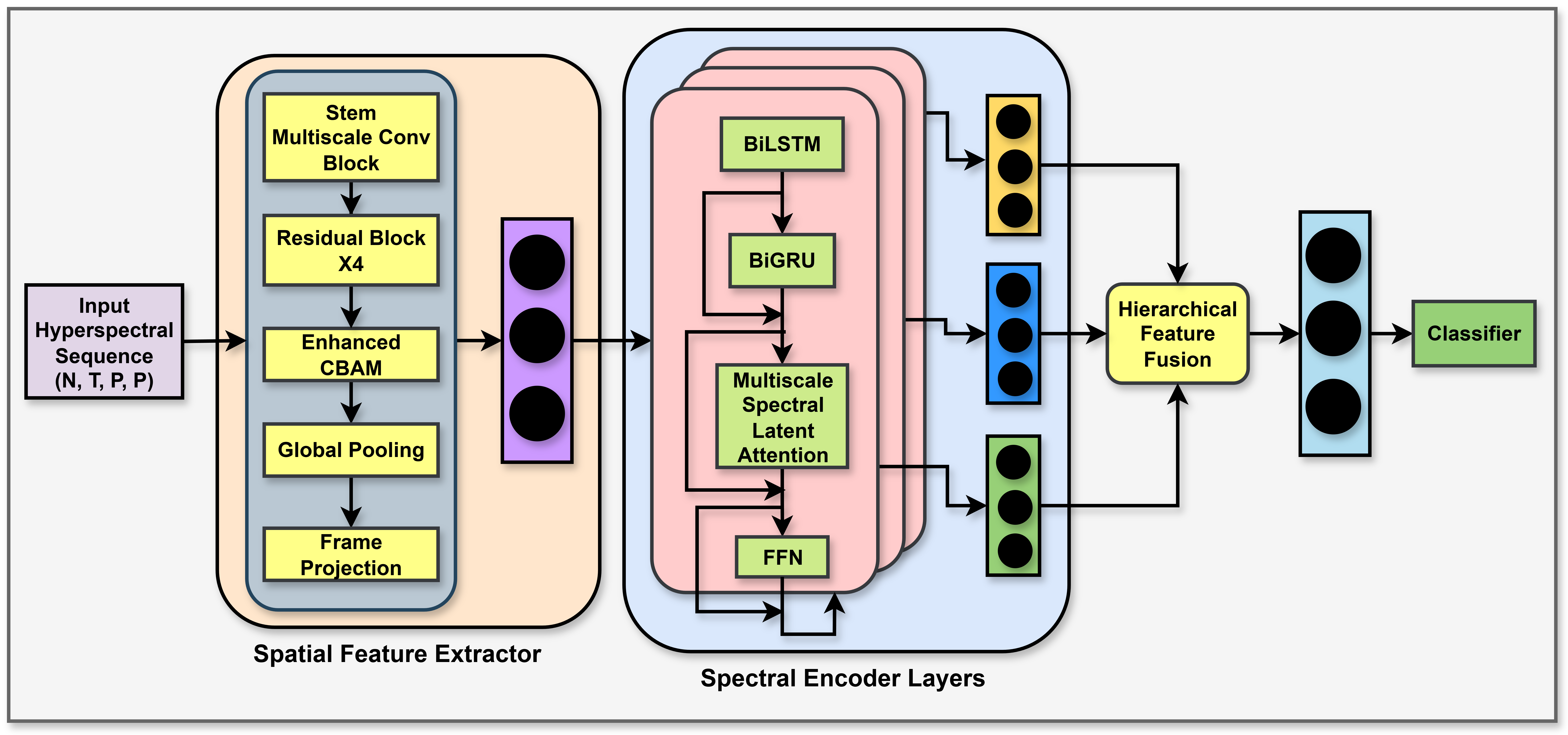} 
\caption{\label{fig:clares} Overview illustration of the proposed CLAReSNet model for HyperSpectral Image Classification}
\end{figure*}

CLAReSNet (Fig.~\ref{fig:clares}) integrates convolutional spatial feature extraction with transformer-style spectral encoding through a latent attention bottleneck. The architecture comprises three stages: spatial feature extraction from individual spectral bands, spectral encoding with multi-scale latent attention, and hierarchical feature fusion for classification.

\subsubsection{\textbf{Spatial Feature Extraction}}
Given input $\mathbf{X} \in \mathbb{R}^{N \times T \times P \times P}$ for a batch of $N$ samples with $T$ spectral bands, each band is processed independently through a multi-scale convolutional stem:
\begin{equation*}
\mathbf{F}_{\text{stem}} = \text{Concat}[\mathcal{C}_1(\mathbf{X}), \mathcal{C}_3(\mathbf{X}), \mathcal{C}_5(\mathbf{X}), \mathcal{C}_7(\mathbf{X})],
\end{equation*}
where $\mathcal{C}_k$ denotes convolution with kernel size $k \times k$. This Inception-inspired design~\cite{ref16} captures multi-resolution spatial patterns from fine textures to broader contexts. The concatenated multi-scale features are further refined by an Enhanced Squeeze-and-Excitation (SE) block~\cite{ref19}, which adaptively recalibrates channel-wise responses using both global average and max pooling statistics:
\begin{equation*}
\mathbf{F}'_{\text{stem}} = \mathbf{F}_{\text{stem}} \otimes \sigma(\text{MLP}([\text{GAP}(\mathbf{F}_{\text{stem}}); \text{GMP}(\mathbf{F}_{\text{stem}})])),
\end{equation*}
where the Multi Layer Perceptron(MLP) consists of two linear layers with reduction ratio 16, GELU~\cite{ref22} activation, and dropout.

Following the stem, four residual blocks are applied, each with progressively increasing dilation rates $d \in \{1, 2, 3, 4\}$:
\begin{equation*}
\mathbf{\hat{F}}_l = \text{BN}(\mathcal{C}_{3,d_l}(\text{Dropout}(\text{GELU}(\text{BN}(\mathcal{C}_3(\mathbf{F}_{l-1})))))),
\end{equation*}
\begin{equation*}
\mathbf{F}_l = \text{GELU}(\mathbf{F}_{l-1} + \text{SE}(\mathbf{\hat{F}}_l)),
\end{equation*}
where $\mathcal{C}_{3,d_l}$ denotes $3 \times 3$ convolution with dilation $d_l$, dilated convolutions expand the receptive field without parameter increase~\cite{ref17}, and the Enhanced SE block adaptively recalibrates channel-wise features before the residual addition. Each residual block applies Enhanced SE attention to the convolutional outputs, enabling the network to emphasize informative channels while suppressing less relevant ones, thereby improving feature discriminability.

Enhanced Convolutional Block Attention Module (CBAM)~\cite{ref18} then applies sequential channel and spatial attention to the residual stack output. Channel attention computes:
\begin{equation*}
\mathbf{F}' = \mathbf{F}_l \otimes \sigma(\text{MLP}([\text{GAP}(\mathbf{F}_l); \text{GMP}(\mathbf{F}_l)])),
\end{equation*}
where GAP and GMP denote global average and max pooling, and $\sigma$ is sigmoid activation. Spatial attention leverages four statistics:
\begin{equation*}
\mathbf{F}'' = \mathbf{F}' \otimes \sigma(\text{Conv}_{7\times7}([\mu(\mathbf{F}'); \max(\mathbf{F}'); \sigma(\mathbf{F}'); \min(\mathbf{F}')])),
\end{equation*}
where operations are performed across channels. Features are aggregated and projected:
\begin{equation*}
\mathbf{e}_t = \mathcal{W}_{\text{proj}}([\text{GAP}(\mathbf{F}''_t); \text{GMP}(\mathbf{F}''_t)]),
\end{equation*}
yielding embeddings $\mathbf{E} = [\mathbf{e}_1, \dots, \mathbf{e}_T] \in \mathbb{R}^{N \times T \times D}$ with dimension $D=256$.

\subsubsection{\textbf{Spectral Positional Encoding}}
Hybrid positional encoding combines sinusoidal and learnable components~\cite{ref15}:
\begin{align*}
\text{PE}(t, 2i) = \sin(t/10000^{2i/D}),\\
\text{PE}(t, 2i+1) = \cos(t/10000^{2i/D}),
\end{align*}
\begin{equation*}
\mathbf{PE}_{\text{hybrid}}(t) = [\mathbf{PE}_{\text{sin}}(t); \mathbf{PE}_{\text{learn}}(t)],
\end{equation*}
where $\mathbf{PE}_{\text{learn}}$ are trainable parameters. Encoded features are 
\begin{equation*}
\tilde{\mathbf{E}} = \mathbf{E} + \mathbf{PE}_{\text{hybrid}}
\end{equation*}

\subsubsection{\textbf{Multi-Scale Latent Attention}}
The core innovation is Multi-Scale Spectral Latent Attention (MSLA), which reduces computational complexity from $\mathcal{O}(T^2D)$ in standard self-attention to approximately $\mathcal{O}(T \log(T) D)$ through adaptive latent bottlenecks~\cite{ref20,ref21}. The number of latent tokens adapts logarithmically:
\begin{equation*}
L'(T) = \left\lfloor L_{\text{base}} \cdot \log_2 \left({\max(T, T_{\text{base}})} / T_{\text{base}} \right) \right\rfloor,
\end{equation*}
\begin{equation*}
L(T) = \min \left( \max \left( L'(T), \, L_{\text{min}} \right), \, L_{\text{max}} \right)
\end{equation*}
with base\_token$(L_{\text{base}})=16$, base\_length$(T_{\text{base}})=16$, min\_tokens$(L_{\text{min}})=8$, max\_tokens($L_{\text{max}})=64$ and $T$ is the input sequence length.

For each scale $s \in \{1, 2, 4\}$, MSLA executes three phases. Encoding compresses information into latents $\mathbf{L} \in \mathbb{R}^{N \times L(T) \times D}$:
\begin{equation*}
\mathbf{Z}^{(s)} = \text{LN}(\mathbf{L} + \text{CrossAttn}(\mathbf{Q}=\mathbf{L}, \mathbf{K}=\tilde{\mathbf{E}}^{(s)}, \mathbf{V}=\tilde{\mathbf{E}}^{(s)})),
\end{equation*}
where $\tilde{\mathbf{E}}^{(s)}$ is the scale-$s$ downsampled input and \text{LN} is layernorm. Processing applies self-attention among latents:
\begin{equation*}
\mathbf{Z}'^{(s)} = \text{LN}(\mathbf{Z}^{(s)} + \text{SelfAttn}(\mathbf{Z}^{(s)})).
\end{equation*}
further processes latents through feed-forward network:
\begin{equation*}
\mathbf{Z}''^{(s)} = \mathbf{Z}'^{(s)} + \text{FFN}(\mathbf{Z}'^{(s)}),
\end{equation*}
where FFN has expansion factor 2 with GELU activation.
Decoding reconstructs outputs via cross-attention:
\begin{equation*}
\mathbf{O}^{(s)} = \text{CrossAttn}\left(\mathbf{Q}=\tilde{\mathbf{E}}^{(s)}, \mathbf{K}=\mathbf{Z}''^{(s)}, \mathbf{V}=\mathbf{Z}''^{(s)}\right)
\end{equation*}
\begin{equation*}
\mathbf{O}^{(s)}_{out} = \text{LayerNorm}\left( \tilde{\mathbf{E}}^{(s)} + \mathbf{O}^{(s)} \right)
\end{equation*}
Multi-head attention~\cite{ref15} with $h=8$ heads computes:
\begin{equation*}
\text{MultiHead}(\mathbf{Q}, \mathbf{K}, \mathbf{V}) = \text{Concat}(\text{head}_1, \dots, \text{head}_h)\mathbf{W}^O,
\end{equation*}
\begin{equation*}
\text{head}_i = \text{softmax}\left(\frac{\mathbf{Q}\mathbf{W}^Q_i(\mathbf{K}\mathbf{W}^K_i)^T}{\sqrt{d_k}}\right)\mathbf{V}\mathbf{W}^V_i.
\end{equation*}
Scale outputs are fused:
\begin{equation*}
\mathbf{F}_{\text{fused}} = \text{LN}(\tilde{\mathbf{E}} + \text{FFN}(\text{Concat}[\mathbf{O}^{(1)}_{out}, \mathbf{O}^{(2)}_{out}, \mathbf{O}^{(4)}_{out}])),
\end{equation*}
where FFN is a two-layer network with GELU activation and expansion factor 2.

\subsubsection{\textbf{Spectral Encoder Layers}}
Three stacked encoder layers combine bidirectional RNNs with MSLA. Each layer $l$ receives the previous layer's output $\mathbf{H}^{(l-1)}$ as input and processes:
\begin{equation*}
\mathbf{H}_{\text{rnn}} = \text{BiGRU}(\text{BiLSTM}(\mathbf{H}^{(l-1)})),
\end{equation*}
\begin{equation*}
\mathbf{H}^{(l)} = \mathbf{H}^{(l-1)} + \text{LN}(\text{FFN}(\text{MSLA}(\mathbf{H}_{\text{rnn}}))),
\end{equation*}
where BiLSTM and BiGRU capture complementary temporal dependencies, FFN has expansion factor 4 with GELU activation, and the residual connection from $\mathbf{H}^{(l-1)}$ facilitates gradient flow through the deep network. The output $\mathbf{H}^{(l)}$ serves as input to the next encoder layer.

\subsubsection{\textbf{Hierarchical Feature Fusion}}
Layer-wise representations are aggregated via cross-attention to leverage multi-level features. Each encoder layer's output sequence is mean pooled to produce a summary vector: $\mathbf{s}^{(l)} = \frac{1}{T}\sum_{t=1}^T \mathbf{H}^{(l)}_t$, and all summaries are stacked as $\mathbf{S} = [\mathbf{s}^{(1)}, \mathbf{s}^{(2)}, \dots, \mathbf{s}^{(n)}]$. The final layer's summary queries the stack through cross-attention:
\begin{equation*}
\mathbf{f}_{\text{final}} = \text{LN}(\mathbf{s}^{(n)} + \text{CrossAttn}(\mathbf{Q}=\mathbf{s}^{(n)}, \mathbf{K}=\mathbf{S}, \mathbf{V}=\mathbf{S})).
\end{equation*}

\subsubsection{\textbf{Classification Head}}
The classification head transforms the fused feature representation into class predictions via progressive dimensionality reduction with differentiated dropout regularization:
\begin{equation*}
\mathbf{z} = \text{GELU}(\mathbf{W}_1 \text{Dropout}_{0.5}(\text{LN}(\mathbf{f}_{\text{final}})) + \mathbf{b}_1),
\end{equation*}
\begin{equation*}
\mathbf{y} = \text{softmax}(\mathbf{W}_2 \text{Dropout}_{0.25}(\mathbf{z}) + \mathbf{b}_2).
\end{equation*}

\section{Implementation Details}

The CLAReSNet architecture is configured with an embedding dimension of 256, 64 base convolutional channels, three stacked spectral encoder layers, and dropout rates of 0.5 in the classification head and 0.1 in internal modules. Enhanced CBAM attention, residual blocks, and multi-scale convolutions are enabled throughout the spatial pathway.

The spatial feature extraction begins with a multi-scale convolutional stem processing each spectral band through parallel convolutions with kernel sizes 1×1, 3×3, 5×5, and 7×7, capturing features from fine textures to broader contexts. The concatenated outputs are refined using an Enhanced Squeeze-and-Excitation (SE) block for improved channel-wise recalibration. This is followed by four residual blocks with progressively increasing dilations [1, 2, 3, 4], each incorporating batch normalization, GELU activation, and Squeeze-and-Excitation channel attention to adaptively modulate intermediate features. The enhanced CBAM module then applies channel attention via parallel global average and max pooling with a shared Multi-Layer Perceptron, followed by spatial attention computed from four channel-wise statistics: mean, maximum, standard deviation, and minimum. The refined features are aggregated through parallel global average and max pooling, concatenated, and projected to the embedding dimension via layer-normalized linear transformation, yielding one feature vector per spectral band.

Spectral positional encoding employs a hybrid approach that combines fixed sinusoidal patterns with learnable parameters, providing both structured inductive bias and adaptive flexibility to capture spectral correlations. The multi-scale spectral latent attention mechanism operates at three temporal scales [1×, 2×, 4×] with adaptively allocated latent tokens (8–64, logarithmically scaled by sequence length). Each scale executes a three-phase attention pipeline: encoding via input-to-latent cross-attention that compresses spectral information into the latent bottleneck, processing through latent-to-latent self-attention and latent feed-forward neural network enabling complex reasoning in compressed space, and decoding via latent-to-output cross-attention reconstructing enriched representations. All attention operations use eight heads with dropout 0.1. Scale outputs are concatenated and fused through a two-layer FFN with GELU activation and residual connections.

Each spectral encoder layer sequentially processes inputs through bidirectional LSTM and GRU networks, capturing long-range dependencies with complementary gating mechanisms. The RNN outputs with residual connections feed into the multi-scale latent attention module, followed by layer normalization and residual connections. A feed-forward network with expansion factor 4, GELU activation, and dropout transforms the attended features, with subsequent residual connection and normalization. Each encoder layer receives the output of the previous layer and, after processing, adds its result back via a residual connection. The residual connections between encoder layers enable gradient flow and hierarchical refinement across the three-layer stack.

Hierarchical feature fusion aggregates representations from all encoder layers via cross-attention. Each layer's output sequence is mean-pooled to produce a summary vector, and these summaries are stacked. The final layer summary queries this stack through cross-attention, dynamically weighting contributions from different depths. The attended output combines with the final layer summary via residual connection and normalization, yielding the final feature representation. The classification head applies layer normalization, dropout (0.5), linear projection to dimension 128, GELU activation, dropout (0.25), and final linear projection to class logits, providing progressive dimensionality reduction with differentiated regularization to prevent overfitting.

\section{Results \& Discussions}

\begin{figure*}[h!]
\centering
\includegraphics[height=8cm]{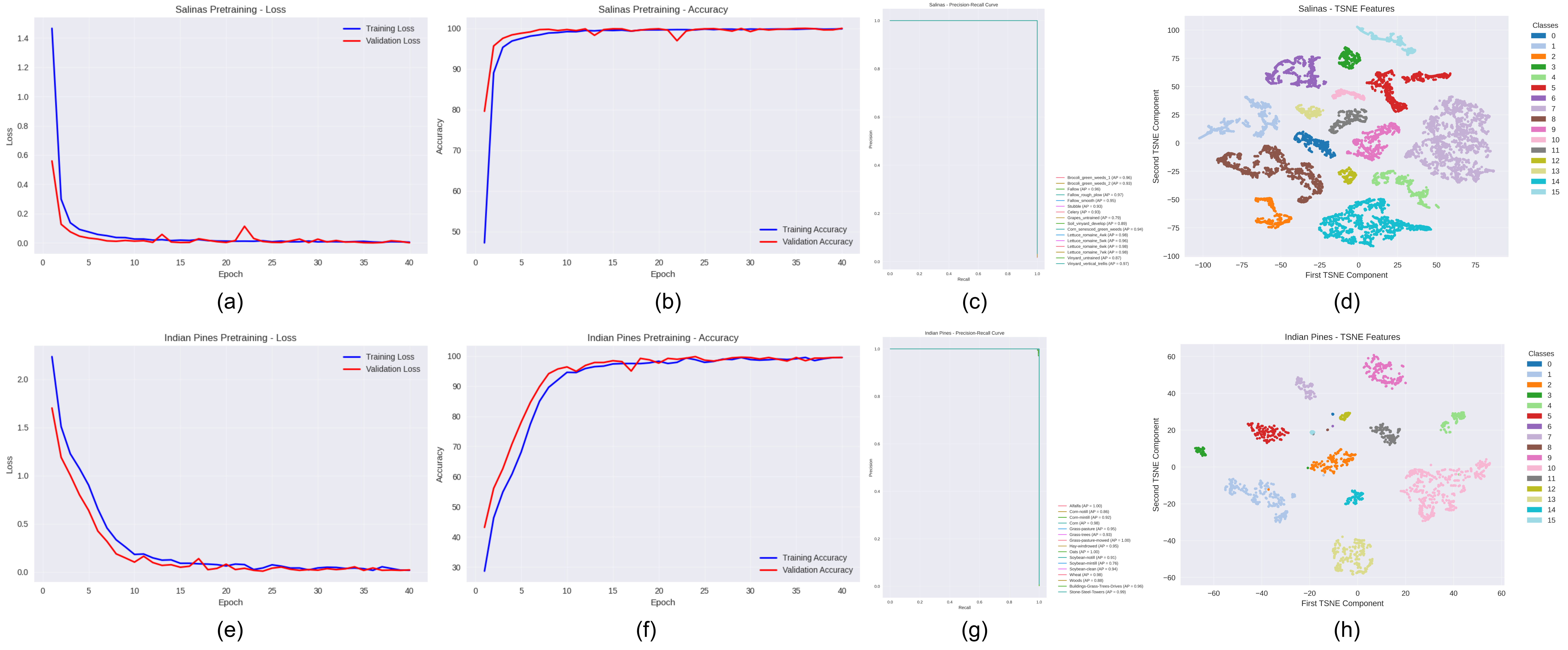} 
\caption{\label{fig:plots} (a) Salinas Loss Curve, (b) Salinas Accuracy Curve, (c) Salinas Precision-Recall Curve, (d) Salinas TSNE Plot, (e) Indian Pines Loss Curve, (f) Indian Pines Accuracy Curve, (g) Indian Pines Precision-Recall Curve, (h) Indian Pines TSNE Plot}
\end{figure*}

\begin{table*}[t]
\caption{Classification Performance on the Indian Pines (IP) Dataset (Mean $\pm$ Std. dev. \%)}
\label{tab:results_ip}
\centering
\renewcommand{\arraystretch}{1.3}
\newcolumntype{C}{>{\centering\arraybackslash}p{2.2cm}}
\begin{tabular}{l c c c c c c}
\hline 
\textbf{Model} & \textbf{OA (\%)} & \textbf{BA (\%)} & \textbf{$\kappa$} & \textbf{MCC} & \textbf{ARI} & \textbf{Avg. Dist.} \\
\hline
Random Forest  & 88.45 $\pm$ 0.52 & 87.12 $\pm$ 0.61 & 0.8651 $\pm$ 0.0078 & 0.8544 $\pm$ 0.0083 & 0.8120 $\pm$ 0.0105 & --- \\
XGBoost        & 92.18 $\pm$ 0.41 & 91.05 $\pm$ 0.49 & 0.9087 $\pm$ 0.0055 & 0.9014 $\pm$ 0.0064 & 0.8803 $\pm$ 0.0098 & --- \\
HybridSN       & 96.53 $\pm$ 0.28 & 95.88 $\pm$ 0.34 & 0.9590 $\pm$ 0.0044 & 0.9551 $\pm$ 0.0048 & 0.9413 $\pm$ 0.0067 & 7.29 $\pm$ 2.10 \\
SSRN           & 97.01 $\pm$ 0.25 & 96.40 $\pm$ 0.30 & 0.9651 $\pm$ 0.003 & 0.9622 $\pm$ 0.0032 & 0.9502 $\pm$ 0.0057 & 11.80 $\pm$ 1.91 \\
SpectralFormer & 73.22 $\pm$ 2.21 & 59.37 $\pm$ 3.38 & 0.6908 $\pm$ 0.0013 & 0.7145 $\pm$ 0.0063 & 0.6935 $\pm$ 0.0017 & 19.55 $\pm$ 1.04 \\
\textbf{CLAReSNet (Ours)} & \textbf{99.71 $\pm$ 0.11} & \textbf{99.78 $\pm$ 0.07} & \textbf{0.9967 $\pm$ 0.0012} & \textbf{0.9967 $\pm$ 0.0004} & \textbf{0.9939 $\pm$ 0.0019} & \textbf{21.25 $\pm$ 1.20} \\
\hline
\end{tabular}
\end{table*}

\begin{table*}[t]
\caption{Classification Performance on the Salinas (SA) Dataset (Mean $\pm$ Std. dev. \%)}
\label{tab:results_sa}
\centering
\renewcommand{\arraystretch}{1.3}
\newcolumntype{C}{>{\centering\arraybackslash}p{2.2cm}}
\begin{tabular}{l c c c c c c}
\hline
\textbf{Model} & \textbf{OA (\%)} & \textbf{BA (\%)} & \textbf{$\kappa$} & \textbf{MCC} & \textbf{ARI} & \textbf{Avg. Dist.} \\
\hline
Random Forest  & 93.81 $\pm$ 0.33 & 94.02 $\pm$ 0.30 & 0.9310 $\pm$ 0.0042 & 0.9263 $\pm$ 0.0055 & 0.9018 $\pm$ 0.0069 & --- \\
XGBoost        & 95.50 $\pm$ 0.27 & 95.61 $\pm$ 0.25 & 0.9500 $\pm$ 0.0030 & 0.9453 $\pm$ 0.0042 & 0.9281 $\pm$ 0.0047 & --- \\
HybridSN       & 98.22 $\pm$ 0.19 & 98.31 $\pm$ 0.21 & 0.9803 $\pm$ 0.0022 & 0.9781 $\pm$ 0.0034 & 0.9712 $\pm$ 0.0032 & 8.10 $\pm$ 1.13 \\
SSRN           & 98.73 $\pm$ 0.15 & 98.80 $\pm$ 0.18 & 0.9855 $\pm$ 0.0024 & 0.9841 $\pm$ 0.0020 & 0.9790 $\pm$ 0.0035 & 12.99 $\pm$ 2.23 \\
SpectralFormer & 89.03 $\pm$ 0.12 & 90.08 $\pm$ 0.15 & 0.8744 $\pm$ 0.0010 & 0.8720 $\pm$ 0.0023 & 0.8549 $\pm$ 0.0029 & 18.33 $\pm$ 1.13 \\
\textbf{CLAReSNet (Ours)} & \textbf{99.96 $\pm$ 0.04} & \textbf{99.98 $\pm$ 0.02} & \textbf{0.9996 $\pm$ 0.0002} & \textbf{0.9996 $\pm$ 0.0002} & \textbf{0.9986 $\pm$ 0.0005} & \textbf{20.98 $\pm$ 1.58} \\
\hline
\end{tabular}
\end{table*}

Experiments on the Indian Pines (IP) and Salinas (SA) hyperspectral benchmark datasets evaluate CLAReSNet against traditional machine learning methods (Random Forest, XGBoost) and state-of-the-art deep learning models (HybridSN, SSRN, SpectralFormer).

On Indian Pines, CLAReSNet achieves an overall accuracy (OA) of 99.71\%, balanced accuracy (BA) of 99.78\%, Cohen's kappa ($\kappa$) of 0.9967, Matthews correlation coefficient (MCC) of 0.9967, and adjusted Rand index (ARI) of 0.9939, outperforming SSRN by 2.70 percentage points in OA. It attains the highest average inter-class distance of 21.25, demonstrating superior feature separability. Despite severe class imbalance (e.g., Oats with fewer than 50 samples), CLAReSNet maintains robust performance across all classes.

On Salinas, CLAReSNet records 99.96\% OA, 99.98\% BA, $\kappa = 0.9996$, and ARI = 0.9986, surpassing SSRN by 1.23 percentage points and achieving near-perfect classification even with high spectral similarity among vegetable classes. The consistently superior performance across both datasets, differing in spatial resolution, class distribution, and complexity, validates CLAReSNet's strong generalization and robustness.

Training curves (Fig.~\ref{fig:plots}) show rapid convergence within $\sim$10 epochs with minimal overfitting, enabled by AdamW optimizer with weight decay, cross-entropy loss, aggressive dropout, and extensive data augmentation. Indian Pines exhibits slightly higher validation variance due to limited samples and imbalance, while Salinas demonstrates stable, smooth convergence. 
The precision-recall curves (Fig.~\ref{fig:plots}) for Indian Pines and Salinas datasets demonstrate CLAReSNet’s strong class-wise discriminative performance, with most classes achieving Average Precision (AP) scores above 0.90 and several reaching perfect precision. This validates the model’s ability to learn well separated feature representations, especially under sparse labels. Challenging classes like Soybean-clean and Grapes-untrained show lower AP due to spectral overlap, but CLAReSNet’s multi-scale latent attention and hierarchical fusion mitigate confusion.

Classification maps (Fig.~\ref{fig:datasets}) reveal spatially coherent predictions with sharp boundaries and minimal salt-and-pepper noise, attributed to 11$\times$11 multi-scale spatial context and enhanced CBAM attention. Uncertainty maps (Fig.~\ref{fig:datasets}) display low entropy in homogeneous regions, with higher uncertainty appropriately localized at class boundaries and mixed pixels, indicating well-calibrated confidence estimates.
t-SNE visualizations of the 256-dimensional learned embeddings (Fig.~\ref{fig:plots}) show highly compact and well-separated clusters for all classes. On Indian Pines, all 16 classes are clearly distinguishable despite minority classes having as few as 20 training samples (e.g., Oats). Salinas exhibits even tighter clustering with virtually no overlap. These results, supported by high inter-class distances (21.25 for IP, 20.98 for SA), confirm that the hierarchical fusion of multi-scale latent attention, bidirectional RNNs, and cross-layer attention mechanisms produces a highly discriminative embedding space with excellent intra-class compactness and inter-class separation.

\section{Conclusion}
In this paper, we introduced the Convolutional Latent Attention Residual Spectral Network (CLAReSNet), a 17.3M-parameter deep hybrid architecture designed to overcome the limitations of using singular 2D or 3D operators. Our approach synergistically fuses rich, localized spatial features from a 2D-CNN stem with a deep spectral encoder capable of modeling long-range sequential dependencies. A key contribution of this work is the \textbf{Multi-Scale Spectral Latent Attention} (MSLA) mechanism. This module, when integrated with recurrent layers, demonstrated a powerful capability for modeling complex, nonlinear spectral relationships at multiple scales, which is critical for differentiating spectrally similar materials. CLAReSNet achieved state-of-the-art (SOTA) accuracy on the Indian Pines and Salinas benchmark datasets, outperforming several established classical machine learning, 2D-CNN, 3D-CNN, and Transformer-based models. This confirms that a parameter-rich hybrid architecture is a highly effective paradigm for HSI analysis. For future work, we plan to investigate the model's scalability and generalization by training it on next-generation, high-resolution data from the PRISMA (PRecursore IperSpettrale della Missione Applicativa) dataset.

\end{document}